\documentclass[11pt]{article}
\usepackage[final]{emnlp2021}
\usepackage{times}
\usepackage{latexsym}
\usepackage{booktabs}
\usepackage{multirow}

\usepackage{microtype}

\usepackage{amssymb,bm,mathtools,color}

\usepackage[normalem]{ulem}
\usepackage{enumitem}

\usepackage{subcaption}
\usepackage{tabularx}
\usepackage{multicol}
\usepackage{multirow}
\usepackage{colortbl}
\usepackage{hhline}
\usepackage{stfloats}
\usepackage{float}

\newcolumntype{Y}{>{\centering\arraybackslash}X}

\usepackage{xcolor}
\definecolor{red}{rgb}{0.74,0.08,0.10}
\definecolor{green}{rgb}{0.26,0.49,0.18}
\definecolor{blue}{rgb}{0.22,0.53,0.75}

\makeatletter
\newcommand\notsotiny{\@setfontsize\notsotiny\@viiipt\@ixpt}
\makeatother

\newcommand{\form}[1]{{\textcolor{magenta!100}{\fontfamily{qag}#1}}}
\newcommand{\controlled}[1]{{\textcolor{blue!100}{\fontfamily{qag}#1}}}

\usepackage{xspace}
\newcommand{\constrainFunction}{\texttt{nextTokens}\xspace}
\newcommand{\scfgrule}[3]{#1 \rightarrow \controlled{#2}\; ,\; \form{#3}}
\newcommand{\scfgruleeq}[3]{{\bf #1} &\rightarrow \texttt{\controlled{#2}}\; ,\; \texttt{\form{#3}}}

\usepackage{listings}

\lstdefinestyle{simple_lst_style}{
    columns=flexible,
    keywordstyle=\color{red},
    numberstyle=\color{gray},
    stringstyle=\color{green},
    basicstyle=\ttfamily\small,
    identifierstyle=\color{black},
    commentstyle=\color{blue},
    breakatwhitespace=false,
    breaklines=true,
    captionpos=b,
    keepspaces=false,
    numbersep=5pt,
    showspaces=false,
    showstringspaces=false,
    showtabs=false,
    tabsize=2,
    frame=single,
}

\lstdefinelanguage{Prompt}{
    morekeywords={Human, Computer},
    sensitive=true,
    morestring=[b]",
}

\title{Constrained Language Models Yield Few-Shot Semantic Parsers}

\author{
  Richard Shin, Christopher H. Lin, Sam Thomson, Charles Chen,\\
  \textbf{Subhro Roy, Emmanouil Antonios Platanios, Adam Pauls,}\\
  \textbf{Dan Klein, Jason Eisner, Benjamin Van Durme} \\
  Microsoft Semantic Machines \\
  \texttt{sminfo@microsoft.com}  
}

\date{}

\begin{document}

\maketitle

\begin{abstract}

We explore the use of large pretrained language models as few-shot semantic parsers.
The goal in semantic parsing is to generate a \emph{structured meaning representation} given a natural language input.
However, language models are trained to generate \emph{natural language}.
To bridge the gap, we use language models to paraphrase inputs into a controlled sublanguage resembling English that can be automatically mapped to a target meaning representation.
Our results demonstrate that with only a small amount of data and very little code to convert into English-like representations, our blueprint for rapidly bootstrapping semantic parsers leads to surprisingly effective performance on multiple community tasks, greatly exceeding baseline methods also trained on the same limited data.

\end{abstract}

\section{Introduction}
\label{sec:introduction}

Large pretrained language models (LMs) like GPT-3~\cite{Brown:2020:gpt-3} have shown increasingly impressive few-shot performance by formulating tasks as text-to-text generation problems~\cite{Raffel:2020:t5,Brown:2020:gpt-3}. Given only a trained LM and a short textual \emph{prompt} that describes and/or exemplifies a task, one can produce surprisingly accurate models for a variety of natural language processing problems.
However, task-specific {\em semantic parsing} does not naturally fit into this paradigm because such parsers typically use custom meaning representations that are unlikely to already exist on the web, let alone exist in large enough quantities to affect the parameters of these LMs. We leverage two key insights to overcome this barrier: (1) since LMs excel at generating natural language, we should formulate semantic parsing as {\em paraphrasing} into a controlled sublanguage~\cite{berant-liang-2014-semantic,Marzoev:20} and (2) 
autoregressive LMs can be efficiently \emph{constrained} to search over only valid paraphrases, so the sublanguage does not need to be learned from scratch.

In particular, following~\citet{berant-liang-2014-semantic}, we envision a developer for some new domain first writing a \emph{synchronous context-free grammar} (SCFG) that defines the space of supported (and well-formed) meaning representations along with canonical natural language constructions that express them.
Such a grammar maps between canonical natural language forms and domain-specific meaning representations, so that a separate LM-based system can focus entirely on mapping an unconstrained utterance $u$ to a canonical (but still natural) form $c$. %
Furthermore, the grammar can be used to constrain this LM-based system so that the LM is only allowed to generate canonical utterances (i.e., utterances that correspond to well-formed meaning representations).

Given such a grammar, an LM, and a handful of examples for priming the LM for the task of interest, our approach immediately yields a working semantic parser. While we do not expect the accuracies of our models to reach state-of-the-art performance when compared to models trained on large amounts of task-specific examples, the ability to rapidly prototype semantic parsers in new domains can be immensely helpful for developers, both by facilitating quick construction of a minimum viable product and by enabling the bootstrapping of new data collection through human-in-the-loop processes~\cite{duan-etal-2016-generating}.

\begin{figure*}

\begin{center}
\includegraphics[width=\textwidth]{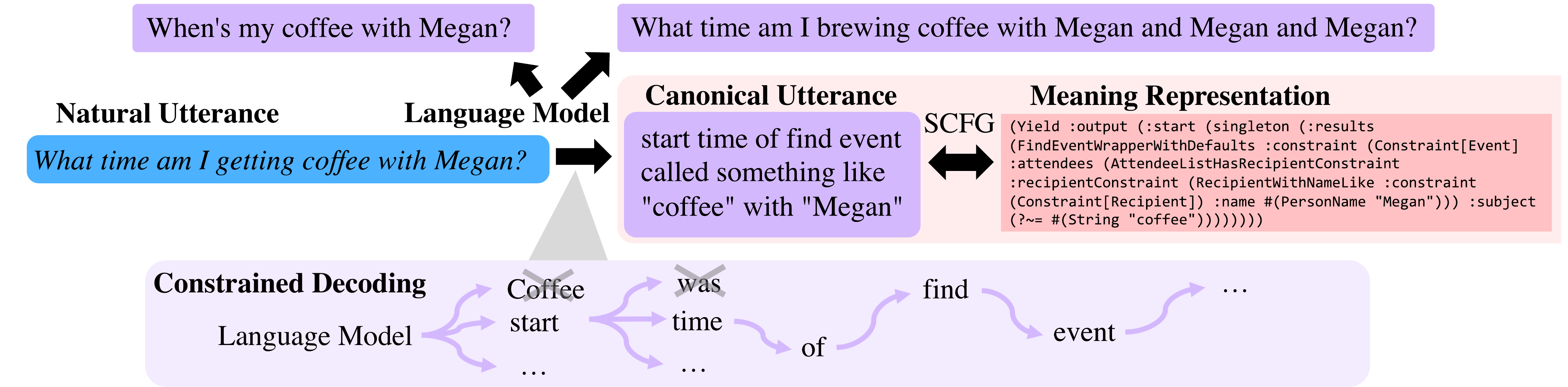}
\end{center}
\caption{%
Our proposed workflow for semantic parsing with a pretrained language model. %
Given a few examples (not shown) and a natural user utterance (blue, italic), a pretrained language model generates paraphrased utterances (purple).
A grammar constrains the search over paraphrases to only canonical utterances, and the highest-scoring canonical paraphrase is mechanically converted to a task-specific meaning representation (pink).
}
\end{figure*}

We report results on the Overnight~\cite{wang-etal-2015-building}, Break~\cite{Wolfson2020Break} and SMCalFlow~\cite{Andreas:2020:dataflow} datasets\footnote{Each of these preexisting datasets uses English inputs, but their output representations vary.} using GPT-2~\cite{Radford:2019:gpt-2}, GPT-3~\cite{Brown:2020:gpt-3}, and BART~\cite{lewis-etal-2020-bart} as the underlying LMs.
Our results demonstrate that our solution:
(1) delivers greater accuracy when LMs target natural language-like representations,
(2) is further improved through the use of explicit decoder constraints; and 
(3) performs surprisingly well with very few examples, suggesting a new frontier for rapidly prototyping semantic parsers. %
The code and grammars developed in this work are publicly available at \url{https://github.com/microsoft/semantic_parsing_with_constrained_lm}.

\section{Background}
\label{sec:background}\label{sec:priming}

\newcommand{\backgroundarea}[1]{\paragraph{#1.}}

\backgroundarea{Autoregressive Language Models}

A language model defines an (estimated) probability distribution over %
sequences of tokens $\bm{w} = w_1, \hdots, w_n$.
Autoregressive LMs factorize this distribution as: 
\begin{align}
p(s) &= \prod_{i=1}^n p(w_i \mid w_1, \hdots, w_{i-1}). \label{eq:lm}
\end{align}
Unlike a cloze model such as BERT \citep{Devlin:2019:bert}, an LM enables text generation, and an autoregressive LM makes it efficient to generate incrementally. %
LMs like GPT-2 \citep{Radford:2019:gpt-2} and GPT-3 \citep{Brown:2020:gpt-3} are trained by maximizing their likelihood on large web corpora. %

It has been shown that autoregressive LMs are powerful at performing tasks not obviously connected to pure language modeling.
For example, \citet{Raffel:2020:t5} showed that an LM was able to extend the prompt {\em ``Translate English to German: That is good.''} with the correct translation {\em ``Das ist gut.''}  \citet{Brown:2020:gpt-3} used ``few-shot'' prompts that included several examples of inputs followed by target outputs, with the actual task input appended at the end.
In both cases, the task defined by the prompt is carried out by asking the language model to generate the subsequent text.
Even without task-specific fine-tuning, this approach has already yielded reasonable results \citep[see e.g.,][]{radford2018improving,Brown:2020:gpt-3,gao2020making}.

This has wide implications, indicating we may be able to carry out various tasks simply by  {\em designing the prompts} that we feed to pretrained LMs, 
removing the expense of training task-specific models.
There already exist multiple approaches to prompt design, like choosing appropriate examples to include in the prompt \citep[e.g.,][]{liu2021makes} or reformulating the prompts into more human-friendly forms \citep[i.e., closer to natural language;][]{schick2020exploiting}.
More related to our work, prompt-guided semantic parsing relates to ideas in example-based machine translation dating back to work by \citet{nagao-1984}, %
that have been recently revisited in the context of semantic parsing with {\em retrieve-and-edit} by \citet{hashimoto-18}.

Fine-tuning can still be used with these models to perform various tasks \citep{li2021prefixtuning,liu2021gpt,schick2020its}.
Although fine-tuning requires additional training, the fine-tuned system can be more efficient at inference time, as it is no longer necessary to select training examples to precede the test input.

\backgroundarea{Semantic Parsing as Paraphrasing}

We adopt the insight from \citet{berant-liang-2014-semantic} that semantic parsing can make use of triples (natural utterance $u$, canonical utterance $c$, meaning representation $m$), where the parser maps $u \mapsto c \mapsto m$.  By design, it is easy to map $c \mapsto m$ and vice-versa. Our innovation is to prompt and constrain an LM so as to make it map $u \mapsto c$.  This approach can exploit newly available large pretrained LMs.

Previous work in parsing as paraphrase has not used generative LMs for the $u \mapsto c$ step.  Rather, it has mapped $u \mapsto c$ by obtaining candidate $c$ values in some way and then \emph{scoring} them according to whether they paraphrase $u$, using a semantic equivalence model that scores $(u,c)$ pairs.
For example, \citet{berant-liang-2014-semantic} mapped from $u$ directly to many candidate meanings $m$, and then evaluated the corresponding canonical utterances $c$ against $u$.
\citet{wang-etal-2015-building} and \citet{Marzoev:20} \emph{generated} candidate $c$ values (along with their meanings $m$) from a grammar of legal canonical utterances, but \emph{incrementally filtered} the bottom-up or top-down generation by scoring the partial candidates against $u$.
Our procedure swaps the roles of the grammar and $u$.  We use $u$ to \emph{generate} the candidate $c$ values by prompting a large LM with $u$, and then \emph{incrementally filter} the left-to-right generation by assessing whether the partial candidates fit the canonical grammar.  This places the LM in the driver's seat.  The large LM that we use for paraphrase \emph{generation} is trained on much more data than the specialized paraphrase \emph{scoring} models used in prior work.\looseness=-1

\backgroundarea{Bootstrapping a Semantic Parser}

One line of prior work on quickly bootstrapping a semantic parser has focused on creating synthetic training examples from a grammar  
developed by hand~\cite{campagna-19,dbpal20,Marzoev:20,campagna-etal-2020-zero} or derived automatically from existing data~\cite{jia-liang-2016-data,yu2020grappa}.
\citet{wang-etal-2015-building} described an approach to bootstrapping that uses a grammar to generate canonical forms, which are paraphrased by crowdworkers to produce training data ``overnight.'' \citet{xu-etal-2020-autoqa} extended this work by generating paraphrases for training data by filtering examples generated from a grammar. %
\looseness=-1

In this paper we take the approach of using the grammar as a \emph{constraint}, with an eye towards enabling bootstrapping through human-in-the-loop semantic parsing,
where humans quickly annotate data by manually correcting parses from an initial prototype \citep{duan-etal-2016-generating,he-etal-2016-human,yao_model-based_2019, elgohary2021nl-edit}.
With this motivation in mind we report accuracy at K, defined as the rate in which an annotator would find the correct parse when selecting among K options.

\section{Approach}
\label{sec:approach}

We propose a method for semantic parsing using large pre-trained LMs that requires little to no task-specific training.
For the prompt-based few-shot setting, %
we use the 175-billion-parameter \mbox{GPT-3} model \citep{Brown:2020:gpt-3} as our LM because at the time of writing %
it was the largest available LM that provided an accessible API.\footnote{\url{https://openai.com/blog/openai-api}.} %
Our goals are to show the approach is good enough to be practical, and %
to confirm our claim that large LMs are better used to generate text that looks more like natural language rather than an artificial programming language.\looseness=-1

Our approach consists of two parts:
(1) LM priming, either through {\it dynamic prompt creation} or {\it fine-tuning}, and
(2) {\em constrained decoding}, ensuring well-formed output  under the target representation.

\paragraph{Dynamic Prompt Creation.}
The prompt we feed to \mbox{GPT-3} is designed so that it contains a small representative set of examples mapping utterances to their desired outputs.
As mentioned in \S\ref{sec:introduction}, we target rapid prototyping and so, for each task that we tackle we assume access to 1,000 or fewer training examples.
Each example is a pair $(u_i, t_i)$ where $u_i$ is an utterance and $t_i$ is the target output for that utterance, specified as either the original meaning representation, $m_i$, or our canonical linguistic representation, $c_i$, which can then be translated to $m_i$. %
Given a test input utterance $u=$ \textit{``how long is the weekly standup''}, for example, a dynamically constructed prompt looks something like:

\vspace{1ex}
\begin{lstlisting}[language=Prompt]
Let's translate what a human user says into what a computer might say.

    Human: when is the weekly standup
 Computer: start time of weekly standup
    Human: what date is the weekly standup
 Computer: date of weekly standup
              ...
    Human: how long is the weekly standup
 Computer:
\end{lstlisting}\vspace{1ex}

\noindent Intuitively, we want the examples used to be similar to the test utterance $u$ %
so  GPT-3 can {\em learn} how to generate the target output based on just the prompt.

We propose to also use GPT-3 for selecting the examples to include in the prompt.
Consider a training example, $(u_i, t_i)$.
We quantify its relevance to the test input $u$ as $p(u \mid u_i)$, computed directly using GPT-3.\footnote{During development we also considered using S-RoBERTa \citep{reimers-gurevych-2019-sentence} and LASER \citep{artetxe-schwenk-2019-massively} for estimating relevance instead of GPT-3, but we did not observe differences significant enough to motivate the additional complexity.%
}
For each test utterance $u$, we sort all training examples by this metric, and construct the prompt from the $P$ most relevant examples.%
Note that the GPT-3 API accepts at most 2,048 tokens (after sub-word tokenization) and thus, if using $P$ exceeds this limit, we reduce $P$ accordingly.
For example, to generate a 40-token %
output we need to limit the prompt size to 2,008 tokens.

\paragraph{Fine-tuning.}
An alternative to few-shot prompting is to fine-tune the LM on each task using \emph{just} the utterance as input.
Since the GPT-3 API available to us does not support fine-tuning, we use the next largest model of the same type, GPT-2 XL.\footnote{We tried using GPT-2 with few-shot prompts and no fine-tuning but the results were sufficiently poor that we did not explore further.}
We also fine-tune BART~\citep{lewis-etal-2020-bart}, a pretrained sequence-to-sequence model with a bidirectional encoder and autoregressive decoder.
As BART is trained to generate sentences given corrupted versions of those sentences, it is perhaps particularly suited for generating paraphrases.

We use the same set of examples to fine-tune that we would otherwise use as candidates for prompt creation,
fine-tuning an LM to do well at mapping utterance $u_i$ to the target output $t_i$; no other examples are included in the prompt.%
When the target is a structured representation, this amounts to sequence-to-sequence semantic parsing.  When the target output is natural language, this might be called text rewriting or sentential paraphrasing.

\label{sec:guided-generation}
\paragraph{Constrained Decoding.}
The input to the LM is a prompt $p$, which always contains the utterance $u$ to be parsed.
In the non--fine-tuned case it is preceded by dynamically constructed examples as described above. Given $p$, we use an LM to generate a continuation $t$ and take this as the output.%
As mentioned in \S\ref{sec:background}, we assume that each target task specifies a way to constrain the generated continuation to guarantee a well-formed output for that task.
Formally, we assume that each task provides a \constrainFunction function which, for any token sequence $s$, %
returns the set of all tokens that can immediately follow $s$ in the target output language.
We then use the LM to produce the output $t$ by extending the prompt $p$ using a length-normalized variant of beam search \citep{Murray:2018:length-normalization,wu2016googles}.
At each step of the search, we filter the set of valid continuations using \constrainFunction.

\section{Case Studies}
\label{sec:case-studies}

In the following sections we present multiple case studies to evaluate our approach.
Each studies a different task and follows the same workflow: a {\it Definition} of the task and the meaning representation it uses; a {\it Framing} of the representation into our proposal, including a description of \constrainFunction; an {\it Experimental Setup} with task-specific details; and {\it Results}, where our experiments evaluate our ability to predict the original meaning representation $m$, either as $u \mapsto m$ or as $u \mapsto c \mapsto m$.

\subsection{Overnight}
\label{sec:overnight}

\paragraph{Definition.}
\citet{wang-etal-2015-building} constructed the Overnight semantic parsing dataset, which contains a total of 13,682 examples across eight different domains exhibiting a variety of linguistic phenomena and semantic structures.
The underlying task aims to map natural language utterances to database queries.
The authors initially generated pairs $(c_i, m_i)$ of canonical utterances and corresponding queries (in the form of Lisp-like S-expressions) using a hand-crafted SCFG.
They then used crowdsourcing to paraphrase each $c_i$ into a more natural-sounding utterance $u_i$.
An example $u_i$ is shown below, followed by the  \controlled{canonical representation $c_i$} and \form{meaning representation $m_i$}:
\begin{center}
\small
\begin{tabular}{p{0.9\columnwidth}}
\emph{which january 2nd meetings is alice attenting [sic]}\\
\midrule
\controlled{meeting whose date is jan 2 and whose attendee is alice}\\
\midrule
\form{(call listValue (call filter}\\
 \form{\quad(call filter (call getProperty}\\
 \form{\quad\quad\quad(call singleton en.meeting) (string !type))}\\
 \form{\quad\quad(string date) (string =) (date 2015 1 2))}\\
 \form{\quad(string attendee) (string =) en.person.alice))}\\
\end{tabular}
\end{center}
\smallskip

\noindent The resulting $(u_i,c_i,m_i)$ triples were used to train a semantic parser that mapped $u \mapsto (c,m)$.

\paragraph{Framing.}
The publicly available release of the Overnight dataset conveniently contains all of the $(c_i, m_i)$ pairs generated by enumerating
SCFG derivation trees up to a certain depth.
For some of these, the natural language paraphrase $u_i$ is also available.
For these, we can directly use $m_i$ as the meaning representation for our setup, and $c_i$ as the canonical representation.
Furthermore, we implement the \constrainFunction function from \S\ref{sec:approach} by building a large trie that contains \emph{all} of the $c_i$ or $m_i$ strings (depending on whether our experimental system is attempting to map $u \mapsto c$ or $u \mapsto m$).%
This trie allows us to quickly look up all the ways in which a valid prefix of a (depth-limited) $c$ or $m$ string can be extended to produce a longer valid prefix. In the case of $m$, it enforces not only syntactic well-formedness but also type safety.

\begin{table*}[t]
\centering
\notsotiny\frenchspacing
\setlength{\tabcolsep}{3pt}
\renewcommand{\arraystretch}{0.9}
\begin{tabularx}{\textwidth}{Xrrrrrrrrr}
\toprule
\textbf{Model} & \textbf{Train} \textit{n} & \textbf{Basketball} & \textbf{Blocks} & \textbf{Calendar} & \textbf{Housing} & \textbf{Publications} & \textbf{Recipes} & \textbf{Restaurants} & \textbf{Social} \\
\midrule
GPT-3 Constrained Canonical & 200 & \textbf{0.859} & \textbf{0.634} & 0.792 & \textbf{0.741} & \textbf{0.776} & \textbf{0.792} & \textbf{0.840} & 0.687 \\
BART$^f$ Constrained Canonical & 200 & 0.847 & 0.581 & \textbf{0.845} & 0.725 & 0.758 & 0.773 & 0.831 & \textbf{0.731} \\
GPT-2$^f$ Constrained Canonical & 200 & 0.836 & 0.549 & 0.804 & 0.640 & 0.752 & 0.787 & 0.762 & 0.726 \\

Cao et al. (2019) & 200 & 0.772 & 0.429 & 0.613 & 0.550 & 0.696 & 0.671 & 0.639 & 0.566 \\
\midrule
\citet{cao-etal-2019-semantic} & 640--3535 & \textbf{0.880} & \textbf{0.652} & \textbf{0.807} & \textbf{0.767} & \textbf{0.807} & \textbf{0.824} & \textbf{0.840} & \textbf{0.838} \\
BERT-LSTM \citep{xu-etal-2020-autoqa} & 640--3535 & 0.875 & 0.624 & 0.798 & 0.704 & 0.764 & 0.759 & 0.828 & 0.819 \\
AutoQA \citep{xu-etal-2020-autoqa} & 0\textsuperscript{$\dagger$} & 0.739 & 0.549 & 0.726 & 0.709 & 0.745 & 0.681 & 0.786 & 0.615 \\
\bottomrule
\end{tabularx}
\caption{Denotation accuracies on Overnight. $^f$ indicates models that have been fine-tuned on the training examples. For results above the line, we use $n = 200$ randomly-sampled training examples; the first three lines are our systems, while for Cao et al. (2019), we ran their training code on the same 200. The results below the line come from prior work using many more training examples. \textsuperscript{$\dagger$}AutoQA was trained on a large set of $>$400,000 synthetic utterances $u$ created from Overnight's canonical utterances by automated paraphrasing.}%
\label{tab:overnight}
\end{table*}

\begin{table*}
\centering
\notsotiny\frenchspacing
\setlength{\tabcolsep}{3pt}
\renewcommand{\arraystretch}{0.9}
\begin{tabularx}{\textwidth}{Xrrrrrrrrr}
\toprule
\textbf{Model} & \textbf{Train} \textit{n} & \textbf{Basketball} & \textbf{Blocks} & \textbf{Calendar} & \textbf{Housing} & \textbf{Publications} & \textbf{Recipes} & \textbf{Restaurants} & \textbf{Social} \\
\midrule
GPT-3 Constrained Canonical & 200 & \textbf{0.80}* & \textbf{0.62}* & \textbf{0.82}* & \textbf{0.71}* & 0.79* & \textbf{0.84}* & \textbf{0.89}* & \textbf{0.72}* \\
GPT-3 Constrained Meaning & 200 & 0.68* & 0.53* & 0.68* & 0.58* & 0.63* & 0.75* & 0.78* & 0.63* \\
GPT-3 Unconstrained Canonical & 200 & 0.76* & 0.46* & 0.68* & 0.56* & 0.58* & 0.74* & 0.74* & 0.55* \\
GPT-3 Unconstrained Meaning & 200 & 0.56* & 0.39* & 0.50* & 0.42* & 0.46* & 0.66* & 0.58* & 0.48* \\
GPT-3 Constrained Canonical & 20 & \textbf{0.80}* & 0.55* & 0.67* & 0.68* & \textbf{0.81}* & 0.60* & 0.76* & 0.67* \\
\midrule
BART$^f$ Constrained Canonical & 200 & \textbf{0.85} & \textbf{0.58} & \textbf{0.85} & 0.73 & 0.76 & \textbf{0.77} & \textbf{0.83} & \textbf{0.73} \\
BART$^f$ Constrained Meaning & 200 & 0.83 & 0.56 & 0.77 & \textbf{0.75} & \textbf{0.79} & 0.76 & 0.81 & 0.69 \\
BART$^f$ Unconstrained Canonical & 200 & 0.83 & 0.56 & 0.80 & 0.67 & 0.72 & 0.75 & 0.81 & 0.65 \\
BART$^f$ Unconstrained Meaning & 200 & 0.82 & 0.55 & 0.76 & 0.71 & 0.77 & 0.73 & 0.80 & 0.63 \\
\bottomrule
\end{tabularx}
\caption{Variations of our method using GPT-3 and BART. ``*'' denotes accuracies computed on a smaller test set randomly sampled from the full set due to the computational cost of using GPT-3.}%
\label{tab:overnight_100}
\end{table*}

\paragraph{Experimental Setup.}
For each domain, we simulate the low-data prototyping regime by using only 200 training examples, randomly selected from the 640--3,535 examples provided in the Overnight training set;
with GPT-3, we also try 20 training examples as a more extreme case.
For each evaluation example, we create the GPT-3 prompt by selecting up to $P=20$ training examples. %
When using constrained decoding, we perform beam search with a beam size of 10.
For unconstrained decoding with GPT-3, we use the API to greedily sample (using a softmax temperature of 0.0) from the prompt until we reach a newline character; we also try beam search with beam size 10, but to save on computation costs, we do so only for the calendar domain.
For parity, we report results using greedy search for unconstrained decoding with models other than GPT-3.\looseness=-1

\begin{figure}[t]
    \centering
    \includegraphics[width=\columnwidth]{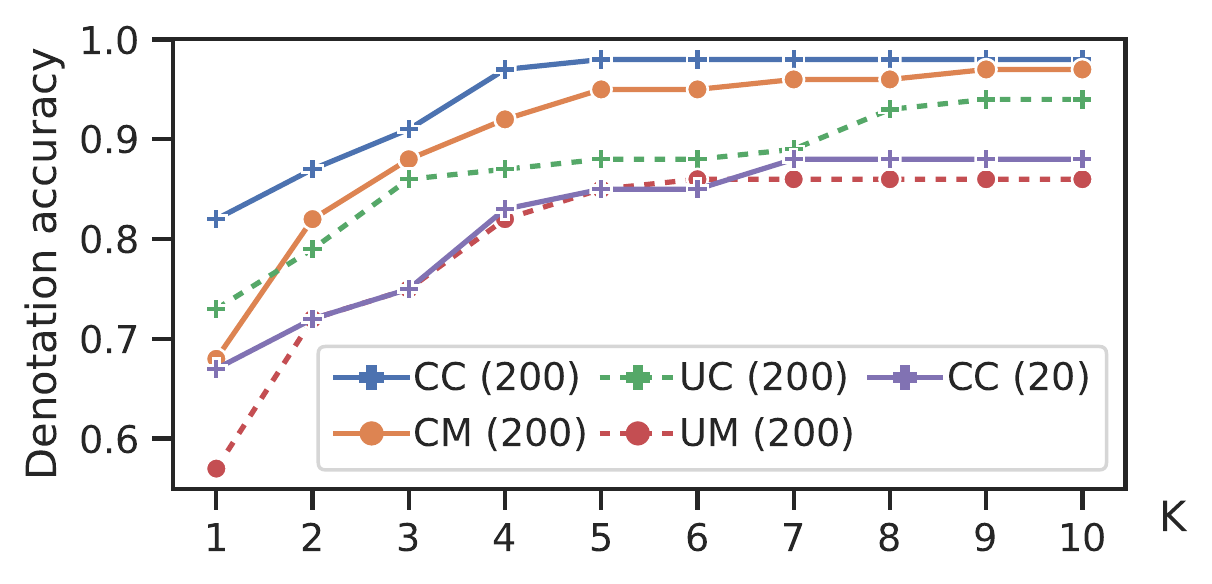}
    \caption{Denotation accuracy @$K$ on the Calendar subdomain of Overnight with GPT-3. Unlike Table~\ref{tab:overnight}, all conditions use beam search; Constrained Canonical (CC), Constrained Meaning (CM), Unconstrained Canonical (UC), and Unconstrained Meaning (UM), using (200) or (20) training examples.}%
    \label{fig:overnight_cal_accuracy}
\end{figure}

\paragraph{Results.}
Table~\ref{tab:overnight} shows our main results on the full test sets in Overnight.
As in prior work we compute the denotation accuracy, checking whether execution of the predicted $m$ against a database returns the gold answer,%
rather than exact match accuracy.
We compare against the current state-of-the-art method from \citet{cao-etal-2019-semantic} also trained on only 200 examples (see Appendix~\ref{sec:caoetal-repro} for details).

Table~\ref{tab:overnight} also includes results using all training examples, from \citet{cao-etal-2019-semantic} and \citet{xu-etal-2020-autoqa}; and AutoQA, which uses only synthetic utterances created by automatically paraphrasing the canonical utterances.
On some of the domains, such as Calendar, Housing, and Restaurants, we obtain similar numbers as the state-of-the-art approach using 7 to 13 times less training data. %

Our method with GPT-3 performs the best among models trained on only 200 examples, approaching the performance of the models trained on all training examples. BART and GPT-2, when fine-tuned on the 200 examples, also perform quite well. BART outperforms GPT-2 despite having fewer parameters, suggesting that its denoising training objective is particularly effective for paraphrasing. Given that fine-tuning was necessary for decent performance for GPT-2, we expect that fine-tuning GPT-3 may improve its performance even further -- when it becomes practical to do so.

Table~\ref{tab:overnight_100} shows that both constrained decoding and the use of English-like canonical utterances rather than Lisp-like logical forms substantially increases the accuracy. This same pattern holds for BART and GPT-2 as well.
Using only 20 training examples generally decreases accuracy by a modest amount, but surprisingly not on all domains.

Figure\@ \ref{fig:overnight_cal_accuracy} shows  accuracy@$K$ on the calendar domain, where the GPT-3 parser is scored as correct on an input if any output in its top $K$ hypotheses is correct.  
The accuracy@5 of Constrained Canonical is 0.98, even though this is only a rapid prototype trained on 200 examples.  %

\subsection{Break}
\label{sec:break}

\paragraph{Definition.}
Break~\citep{Wolfson2020Break} pairs natural language questions with programs in the question decomposition meaning representation (QDMR).
Each program is a sequence of database queries in a controlled natural language, where each query can use the return values of previous queries.
The utterances $u$ are questions sampled from many existing language understanding datasets.\footnote{Break covers semantic parsing~\cite{price-1990-evaluation,zelle:aaai96,Li-nalir:2014,yu-etal-2018-spider}, reading comprehension~\cite{talmor-berant-2018-web,yang-etal-2018-hotpotqa,dua-etal-2019-drop,abujabal-etal-2019-comqa}, and visual question answering~\cite{Johnson_2017_CVPR,suhr-etal-2019-corpus}.}
Crowdworkers \emph{decomposed} each  question $u_i$ into a sequence $m_i$ of queries specified as strings.
The string of each step was restricted to:
(i) words and their inflections appearing in the questions,
(ii) 66 pre-defined function words (e.g., \emph{``if''}, \emph{``on''}, or \emph{``for each''}), and
(iii) tokens that refer to results from the previous step.
This resulted in 44,321 train, 7,760 development, and 8,069 test examples.
An example is shown below, including our \controlled{canonical representation} (defined next) and the QD \form{meaning representation}:

\begin{center}
\small
\begin{tabular}{p{0.9\columnwidth}}
\emph{What color are a majority of the objects?}\\
\midrule
\controlled{(colors of (objects)) where (number of (objects for each (colors of (objects))) is highest)}\\
\midrule
\form{1. objects}\\
\form{2. colors of \#1}\\
\form{3. number of \#1 for each \#2}\\
\form{4. \#2 where \#3 is highest}\\
\end{tabular}
\end{center}

\paragraph{Framing.}
For our canonical representation $c_i$, we mechanically and invertibly map the QDMR $m_i$ into a single-line format that more closely resembles a detailed English request, as illustrated above.
We implement \constrainFunction by restricting the allowed tokens to:
(i) words or their inflections that appear in the questions,
(ii) the pre-defined set of function words, and
(iii) opening and closing parentheses.
A string is considered valid if its tokens belong to one of these three categories, and any parentheses used are balanced.

\paragraph{Experimental Setup.}
The Break leaderboard\footnote{\url{https://leaderboard.allenai.org/break}.}  reports four metrics, with a focus on normalized exact match accuracy (NEM), defined as exact match accuracy after QDMR canonicalization.
All four metrics followed consistent relative trends in our experiments; we focus on NEM for brevity and clarity.
We sampled $n \in \{ 25, 100, 200, 1000 \}$ items uniformly at random from the training set to simulate varying amounts of data in the low-data, rapid prototyping regime.
For each evaluation example, we create the prompt by selecting up to $P = 20$ of the $n$ available training examples.

\begin{table}
\centering
\notsotiny\frenchspacing
\setlength{\tabcolsep}{3pt}
\renewcommand{\arraystretch}{0.9}
\begin{tabularx}{\columnwidth}{Xrl}
\toprule
{\bf Model} & \textbf{Train} \textit{n} & {\bf nem}   \\ %
\midrule
Wolfson et al. & 44,321&0.42   \\
Coleman \& Reneau & 44,321 & 0.42	\\
\midrule
GPT-3 Constrained Canonical &1,000 & 0.32* \\
GPT-3 Constrained Canonical &100 & 0.24* \\
GPT-3 Constrained Canonical &25 & 0.20*\\
\midrule
GPT-3 Constrained Canonical &200 & 0.31* \\
GPT-3 Constrained Meaning & 200 & 0.24* \\
GPT-3 Unconstrained Canonical & 200 & 0.20* \\
GPT-3 Unconstrained Meaning & 200 & 0.17* \\
\midrule
GPT-3 Constrained Canonical & 200 & 0.24 \\
BART$^f$ Constrained Canonical & 200 & 0.22 \\
BART$^f$ Constrained Meaning & 200 & 0.22 \\
BART$^f$ Unconstrained Canonical & 200 & 0.18 \\
BART$^f$ Unconstrained Meaning & 200 & 0.19 \\
\bottomrule
\end{tabularx}
\caption{NEM accuracy on the Break dataset, where $n$ is the number of training examples used in each case. 
Entries drawn from the task leaderboard are included as reference points. 
``*'' denotes accuracies on a random sample on validation. $^f$ indicates fine-tuned models.
}%
\label{tbl:break}
\end{table}

\begin{figure}
    \centering
    \includegraphics[width=0.9\columnwidth]{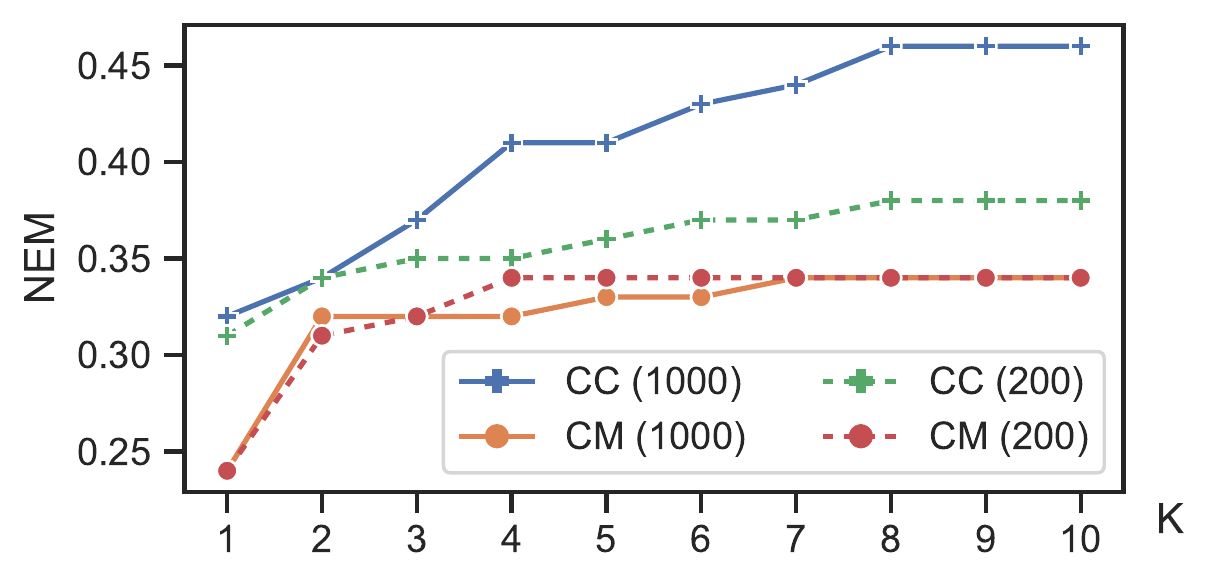}
    \caption{NEM @$K$ on a sample from Break for GPT-3 Constrained Canonical (CC) and Constrained Meaning (CM) with 200 training examples.}
    \label{fig:break_nem}
\end{figure}

\paragraph{Results.}
Table~\ref{tbl:break} shows the results.
Similar to the first case study (\S\ref{sec:overnight}), we observe that our Constrained Canonical approach obtains competitive accuracy despite using relatively few training examples.
We can see that the canonical representation is easier to predict than the meaning representation, even though QDMR was already designed to be more natural than the original representations of the various Break datasets.
We also see that constrained decoding results in further improvements, leading to gains of 7--11\% in absolute accuracy. All our methods outperform a standard seq2seq %
baseline (BART Unconstrained Meaning) trained to predict the meaning representation on the same number of training examples.
For constrained decoding of canonical utterances, we see steady improvements as we increase the number of training examples from 25 up to 1,000.
Figure~\ref{fig:break_nem} illustrates accuracy over the top-$K$ predictions of the model, with results consistent with %
\S\ref{sec:overnight}.\looseness=-1

\subsection{SMCalFlow}
\label{sec:smcalflow}

\paragraph{Definition.} SMCalFlow~\cite{Andreas:2020:dataflow} is a large dataset for task-oriented dialogue, spanning  the domains
of events, weather, places, and people. User utterances $u$ in SMCalFlow are paired with
rich executable dataflow programs $m$ featuring API calls, function composition, complex constraints, 
and references to the programs from previous dialogue turns. The dataset contains 133,821 training examples $(u_i,m_i)$. Examples have a history of previous dialogue turns, which we ignore here in order to align our approach with the previous sections.\footnote{Ignoring dialogue history hurts performance  relative to prior work: history could be incorporated into a prompt in future work that strives for state of the art.}
The following example shows the \controlled{canonical representation} (defined next) and the \form{meaning representation}:\looseness=-1

\begin{center}
\small
\begin{tabular}{p{0.95\columnwidth}}
\emph{What did I set as my response status for the team meeting?}\\
\midrule
\controlled{my response status of find event called something like ``team meeting''}\\
\midrule
\form{(Yield :output}\\
\form{\quad(:responseStatus (singleton (:results}\\
\form{\quad\quad(FindEventWrapperWithDefaults}\\
\form{\quad\quad\quad:constraint (Constraint[Event]}\\
\form{\quad\quad\quad\quad:subject (?$\sim$= \#(String ``team meeting''))))))))}\\
\end{tabular}
\end{center}

\paragraph{Framing.} Unlike the previous datasets, SMCalFlow does not come with a grammar.
For such a complex dataset, writing a grammar post-hoc that can produce fluent, natural English is challenging.  At the same time, SMCalFlow is representative of the rich semantic parsing tasks our proposal is meant to help rapidly prototype hence its inclusion.%

In order to map between $m$ and a canonical utterance $c$, we built an SCFG over $(c,m')$ pairs, where $m'$ is a transformed intermediate representation that is more SCFG-friendly than $m$ (see Appendix~\ref{sec:scfg-calflow} for details).
While our transformation and SCFG allow us to map $m \mapsto m' \mapsto c$ deterministically (to construct training examples $(u_i,c_i)$ for the prompt), some simple guessing models are required in the reverse direction $c \mapsto m' \mapsto m$ (to convert GPT-3's linguistic output to the desired SMCalFlow representation), since our canonical utterances $c$ are occasionally ambiguous and since $m'$ omits some information about coreferent nodes. 

From this SCFG, we extract two CFGs that define the well-formed sequences $c$ and $m'$, respectively.
As we generate a prefix from left to right, we incrementally parse it using Earley's algorithm \cite{earley70}.  \constrainFunction inspects the state of the incremental parser to return precisely the set of next tokens that are allowed by the CFG.\footnote{%
To robustly handle tokenization mismatches between the pretrained LM and grammar, we effectively transform the grammar such that terminals are single characters. 
Details in Appendix~\ref{sec:scfg-calflow}.
}

\paragraph{Experimental Setup.} We gather $300$ training examples from the SMCalFlow training set through stratified sampling (see Appendix~\ref{sec:stratified}) to simulate a scenario where examples of different kinds are written by a domain developer in the course of developing annotation guidelines.  We also uniformly sample a set of $100$ examples and use stratified sampling for a set of $150$ examples from the SMCalFlow validation set to assist in grammar development and hyperparameter tuning. We use a beam size of $10$.  For some GPT-3 experiments, we uniformly sample an evaluation set of $200$ examples from the SMCalFlow validation set.

\begin{figure}
    \centering
    \includegraphics[width=\columnwidth]{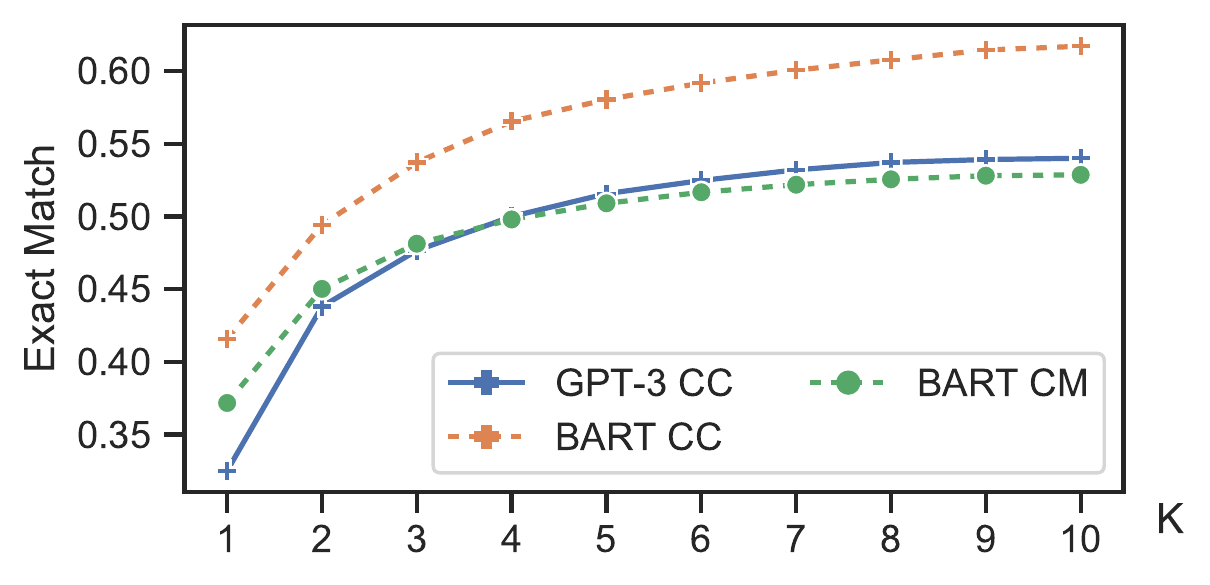}
    \caption{Accuracy@$K$ of three of our models on SMCalFlow.}
    \label{fig:calflow_accuracy}
\end{figure}

\paragraph{Results.} Results are shown in Table \ref{tab:calflow-results} and Figure~\ref{fig:calflow_accuracy}. Note that we always evaluate on the original meaning representation. We find similar relative differences as in previous tasks: targeting a more natural representation and constraining the decoding improves results.
Our methods also significantly outperforms a standard sequence to sequence baseline (BART Unconstrained Meaning) trained to predict meaning representations.

\begin{table}[t]
    \centering
    \notsotiny\frenchspacing
    \setlength{\tabcolsep}{3pt}
    \renewcommand{\arraystretch}{0.9}
    \begin{tabularx}{\columnwidth}{Xrr}
        \toprule  
        {\bf Model} & \textbf{Train} \textit{n} & {\bf Accuracy} \\%
         \midrule
         \citet{Andreas:2020:dataflow} & 133,821 & 0.73 \\
         \midrule
        GPT-3 Constrained Canonical & 300 & 0.33* \\ %
        GPT-3 Constrained Meaning & 300 & 0.25* \\ %
        GPT-3 Unconstrained Canonical & 300 & 0.26* \\ %
        GPT-3 Unconstrained Meaning & 300 & 0.20* \\ %
        \midrule
        GPT-3 Constrained Canonical & 300 & 0.32 \\ %
        BART$^f$ Constrained Canonical & 300 & 0.42 \\
        BART$^f$ Constrained Meaning & 300 & 0.37 \\
        BART$^f$ Unconstrained Canonical & 300 & 0.40 \\
        BART$^f$ Unconstrained Meaning & 300 & 0.30 \\

    \bottomrule
    \end{tabularx}%
    \caption{Performance on SMCalFlow. ``*'' indicates evaluation on a random sample of validation. $^f$ are finetuned models.}
    \label{tab:calflow-results}
\end{table}

\begin{figure}
    \centering
    \includegraphics[width=0.9\columnwidth]{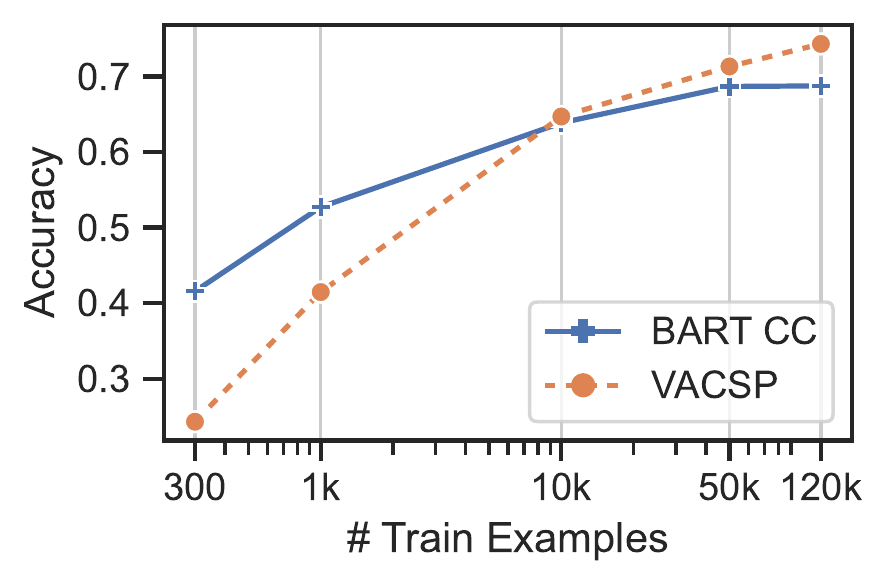}
    \caption{Accuracy of our best model on SMCalFlow at different training set sizes, compared to the recent state of the art model by \newcite{platanios-etal-2021-value}.} %
    \label{fig:data_complexity}
\end{figure}

\paragraph{More Data.} To analyze the effect of training data size, we also evaluate our BART Constrained Canonical
model on stratified training set sizes of 1k, 10k, 50k and on the full SMCalFlow training set ($\approx$120k). For comparison, we also consider the current state-of-the-art model on SMCalFlow  \citep[VACSP;][]{platanios-etal-2021-value} in the same settings. Figure \ref{fig:data_complexity} shows the results. Recall that for all experiments we do not use any dialogue context and so the performance of VACSP is lower than 
the performance reported by \citet{platanios-etal-2021-value}.

Our proposed method outperforms VACSP in low data regimes, further supporting the intuitions behind our approach.  With more training data the benefits of constrained decoding and an initialized decoder become less important.  Future work could assess the relative impact of sampling examples from the grammar for use in pretraining a model such as VACSP, contrasting with using the same grammar as constraints on paraphrase decoding.

\section{Discussion} %

Empirically, we demonstrated that (i) constrained decoding is better than unconstrained and (ii) controlled natural languages are better than meaning representations when used with a pre-trained LM. 
The benefit of (i) is readily observable because unconstrained decoding can produce not-quite-correct answers. For example, GPT-3 constrained decoding maps
the Overnight example \textit{show me any meetings labeled as important which are also three hours long} to the correct canonical utterance \controlled{meeting that is important and whose length is three hours}, whereas unconstrained decoding yields the non-canonical utterance
\controlled{meeting whose label is important and whose length is three hours}.

The effect of (ii) is harder to isolate, though we found some suggestive examples, e.g., for the input utterance \textit{meetings that are not attended by alice}, our method led to the correct \controlled{meeting whose attendee is not alice}.  In contrast, constrained prediction of the meaning representation dropped the negation (using \texttt{=} instead \texttt{!=}), producing the meaning representation for \controlled{meeting whose attendee is alice and is important}.
We speculate that constrained GPT-3 was more willing to preserve the input word \textit{not} than to produce \texttt{!=}.   More impressively, in Break, our method correctly interpreted the novel bigram \textit{as many}, mapping \textit{Are there as many matte objects as metallic objects?} to \controlled{((number of (matte objects)) is same as (number of (metallic objects))}.  In contrast, constrained prediction of the QDMR led to the wrong predicate, whose canonical utterance would be \controlled{((number of (matte objects)) is higher than (number of (metallic objects))}.

\section{Further Related Work}
\label{sec:related-work}

Motivated by tasks where a user requires certain phrases to be present or absent in the output of a text generation system, researchers have explored increasingly more efficient approaches to restricting valid paths in beam search such that they satisfy externally provided constraints~\citep[e.g.,][]{hokamp-liu-2017-lexically,anderson-etal-2017-guided,post-vilar-2018-fast,hu-etal-2019-improved}. %
\emph{Grammar-constrained} decoding restricts some or all of a successful transduction path to result in a sequence parseable under a grammar.
Such techniques were used in task-oriented speech recognition systems \citep{moore-etal-1997-commandtalk},\footnote{Prior to recent advances it was believed that {\it ``practical application of 
speech recognition technology requires a vocabulary and grammar tailored to the particular application, since for high accuracy the recognizer must be restricted as to what sequences of words it will consider''} -- Moore et al. } %
where it was assumed a user knew the precise way to phrase commands.  In contemporary settings we retain the notion of a parser supporting task-specific features, where we would like to enjoy the benefits of a grammar in terms of laying out prescribed functionality but without constraining the user's linguistic forms.
Constraining neural semantic parsing decoders has been explored by
\citet{yin-neubig-2017-syntactic} and \citet{krishnamurthy-etal-2017-neural}, among others, for generating structured forms rather than paraphrases. \citet{herzig2021unlocking} predict intermediate semantic representations with stronger structural correspondence to natural language than $m$, replacing the role of $c$ in our approach with a modified meaning representation $m'$. %

Like the closely related problem of machine translation \citep{wong-mooney-2006-learning,andreas-etal-2013-semantic}, semantic parsing has recently been driven by encoder-decoder neural architectures  \cite[starting with][]{dong-lapata-2016-language,jia-liang-2016-data,kocisky-etal-2016-semantic}. %
More recently, \citet{chen-etal-2020-low} used pre-trained LMs, including BART, to initialize both the encoder and the decoder of a semantic parser. 
In concurrent work, \citet{desai2021lowresource} reports gains on \citet{chen-etal-2020-low} by  modifying a target representation to be more natural language-like.
We argue that LMs are better suited for generating {\em natural language} directly rather than task-specific meaning representations,
using experiments designed to contrast the proficiency of LMs on these two output modalities.

Finally, \newcite{wu-etal-2021-paraphrasing} concurrently proposed a similar solution to our own.  We independently confirm positive results on Overnight, with new studies on Break and SMCalFlow.  In contrast to their primary focus on the unsupervised setting, our experiments were largely concerned with the few-shot scenario.  We consider it reasonable to expect small hundreds of examples from a domain expert when building a real world parser, and our results suggest that this obviates the concerns of Wu et al. on initially tuning a paraphrase model beyond what current off-the-shelf pretraining methods provide.

\section{Conclusion}
\label{sec:conclusion}

We wish to rapidly develop semantic parsers in new domains.  To this end, we have demonstrated that constrained decoding of powerful language models can enable the paraphrasing of user utterances into a controlled sublanguage, which may then be mapped to a task-specific representation.  With small hundreds of examples we are able to quickly bootstrap models for a variety of datasets, enabling future work that explores human in the loop interactions for iterative model refinement.

\bibliography{references}
\bibliographystyle{acl_natbib}

\clearpage
\appendix

\section{SMCalFlow SCFG}
\label{sec:scfg-calflow}
We use a synchronous context-free grammar~(SCFG) to convert between SMCalFlow meaning representations $m$ and canonical English representations $c$. Mapping $m \mapsto c$ is necessary in order to convert the SMCalFlow dataset into prompt examples $(u_i,c_i)$, while mapping $c \mapsto m$ is necessary to convert the predicted canonical English paraphrases back into a target meaning representation. In this section, we review SCFGs and discuss in general how they can be used to map between canonical utterances and meaning representations.  We  describe specific issues that arose in the case of SMCalFlow, and how we handled them.  These techniques may also be useful in other domains.

\subsection{Context Free Grammars}
A context free grammar (CFG) is a 4-tuple $(V, \Sigma, R, v_0)$ where $V$ is a set of nonterminal symbols, $\Sigma$ is a set of terminal symbols, $R = \{V \times (V \cup \Sigma)^*\}$ is a set of rules, and $v_0\in V$ is the starting nonterminal. A CFG is specified by writing a list of rules that \emph{expand} a nonterminal $v \in V$ into a string of nonterminals and terminals, 
\[
v \rightarrow \sigma_0v_1\sigma_1\cdots v_n\sigma_n,
\]
 where $v_i \in V^*, \sigma_i \in \Sigma$.
The language $L$ defined by a CFG consists of all strings that can be generated by using the rules to recursively expand nonterminals starting from the start nonterminal $v_0$ until there are no nonterminals left.
A string $s \in L$ can be parsed into one or more parse trees, which describe the expansions that could have been used to generate $s$. A string is \emph{ambiguous} if there is more than one possible parse for it, and a grammar is ambiguous if any string in its language is ambiguous.
Grammars that attempt to cover natural language tend to be highly ambiguous, but in our setting an unambiguous grammar is preferable.

An SCFG can be thought of as two CFGs that share nonterminals, but have their own set of terminals. Instead of specifying a single expansion, each rule specifies two expansions, a \emph{source} and a \emph{target} expansion, which are synchronized by using the same nonterminals: 
\[
\scfgrule{v}{\sigma_0 v_1 \sigma_1 \cdots v_n \sigma_n}{\tau_0 v_1 \tau_2 \cdots v_n \tau_n}
\]
The two expansions must use the same $n$  nonterminals, although the form above may be generalized to allow these nonterminals to appear in different orders in the source and target expansions. The set of rules and their source expansions defines a CFG and a language $C$, and the set of rules and their target expansions defines a CFG and a language $M$. Because each expansion's nonterminals are the same in any given rule, given an SCFG and a string $c \in C$, we can parse $c$ to obtain a parse tree, and then use this parse tree to generate its corresponding string $m \in M$. While one set of expansions is termed the source and the other the target, we can also reverse the process and translate a string $m \in M$ to a string $c \in C$.
It is this ability to pair two languages together that we use to map between canonical and meaning representations. 

\subsection{SCFG for Semantic Parsing}

Now suppose we have some semantic parsing domain with a set $F$ of functions.  Each function $f \in F$ has a type signature $\form{f(a_1^f: T_1^f, \hdots, a_n^f: T_n^f) \rightarrow T^f}$, where $T_f$ is the return type and $a_i^f$ and $T_i^f$ are the name and type of the $i$\textsuperscript{th} argument.  For simplicity, we treat constants of the domain as 0-ary functions, writing them without the parentheses. 

In the case of SMCalFlow, we had to reconstruct the type signatures for the functions in the dataset, as they were not provided with the dataset release.

For each function, we specify a corresponding English template $E(f) = \controlled{\sigma^f_0 a^f_1 \sigma^f_1 \cdots a^f_n \sigma^f_n}$, where each $\sigma^f_i$ is a possibly empty\footnote{For example, in SMCalFlow, our template for the function \texttt{\form{Execute(\$intension)}} is simply \texttt{\controlled{\$intension}}.} string of English text.  Again, we may generalize to allow the $a_i$ to be ordered differently in $E(f)$ than in $f$.  

We can define an SCFG that maps between programs and English for this domain by writing down the rules
\[
\scfgrule{T^f}{\sigma^f_0 T^f_1 \sigma^f_1 \cdots T^f_n \sigma^f_n}{f(T_1^f,\hdots, T_n^f)}
\]
for all $f \in F$.  Let $\mathcal{T}$ denote the set of types.

For example, consider a toy domain where we can buy colored shapes. We have types $\mathcal{T} = \{{\bf Command}, {\bf CShape}, {\bf Shape}\}$, and functions for returning shapes, coloring those shapes, and buying the colored shapes: 
\begin{align*}
F = \{ 
\form{\texttt{buy(\$o:~CShape)}} &\rightarrow \textbf{Command},\\
\form{\texttt{toRed(\$s:~Shape)}} &\rightarrow \textbf{CShape},\\
\form{\texttt{toGreen(\$s:~Shape)}} &\rightarrow \textbf{CShape},\\
\form{\texttt{square}} &\rightarrow \textbf{Shape},\\
\form{\texttt{triangle}} &\rightarrow \textbf{Shape}
\}
\end{align*}
We could write English templates:
\begin{align*}
E(\texttt{buy}) &= 
\controlled{\texttt{Buy a \$o}}\\
E(\texttt{toRed}) &= \controlled{\texttt{red \$s}}\\
E(\texttt{toGreen}) &= \controlled{\texttt{green \$s}}\\
E(\texttt{square}) &= \controlled{\texttt{box}}\\
E(\texttt{triangle}) &= \controlled{\texttt{triangle}} 
\end{align*}
The resulting SCFG for our toy domain would be:
\begin{small}
\setcounter{equation}{0}
\begin{eqnarray}
&\scfgruleeq{Command}{Buy a {\bf CShape}}{buy({\bf CShape})}\\
&\scfgruleeq{CShape}{red {\bf Shape}}{toRed({\bf Shape})} \\
&\scfgruleeq{CShape}{green {\bf Shape}}{toGreen({\bf Shape})} \\
&\scfgruleeq{Shape}{box}{square} \\
&\scfgruleeq{Shape}{triangle}{triangle} 
\end{eqnarray}
\end{small}
where we have bolded the nonterminals. Now given a canonical English utterance like \texttt{\controlled{Buy a green box}}, we can parse it to produce the parse tree \texttt{(1 (3 (4))}, which we can then use to generate the program \texttt{\form{buy(toGreen(square))}}.

\subsection{Ambiguity}

Ideally, the mappings $c \mapsto m$ and $m \mapsto c$ would be 1-1 mappings, but an SCFG does not guarantee this.
In the case of our SMCalflow SCFG, each meaning representation does have only a single parse---as one would expect for  code in a formal language---so $m \mapsto c$ is deterministic.  Unfortunately, a canonical utterance $c$ may have multiple parses, leading to different meanings $m$.

The first reason for ambiguity arises directly from the ambiguity of English. For example, does \controlled{Create a meeting after the meeting with Bob} mean ``After the meeting, create a meeting with Bob'' or ``After the meeting with Bob, create a meeting''? While one could attempt to wordsmith the templates to eliminate this kind of ambiguity, doing so can quickly become unscalable for large domains.

The other reason that ambiguity occurs is that we allow templates to not contain any English literals, allowing a parser to loop arbitrarily many times on a nonterminal. 
For example, consider the templates for the type coercion functions \texttt{\form{toRecipient(\$person)}} and \texttt{\form{personFromRecipient(\$recipient)}}. Since these functions 
coerce types in a way that English leaves implicit, our English templates for these two functions do not contain any English literals. This leads to SCFG rules like
\begin{eqnarray*}
  &\scfgruleeq{Recipient}{{\bf Person}}{}\\
  & & \texttt{\form{toRecipient({\bf Person})}}\\
    &\scfgruleeq{Person}{{\bf Recipient}}{}\\
  & & \texttt{\form{personFromRecipient({\bf Recipient})}}\\
&\scfgruleeq{Person}{Bob}{Bob}
\end{eqnarray*}
In this case, given the canonical utterance \controlled{Bob}, one can repeatedly apply the first two rules any number of times, producing infinitely many parses.

We could solve both problems by weighting the rules in the grammar, and picking the lowest-weight parse. Since data is available, we can also train a model to predict the correct parse. However, we find that in practice, (1) limiting the allowable recursion in the grammar so that the grammar can only produce a finite number of parses and then (2) using some heuristic rules to pick among those finite set of parses, is both simple and works well.

To limit the recursion in the grammar, we first define a graph induced by the SCFG, where nodes represent nonterminals and a directed edge from node $n_s$ to $n_d$ represents usage of the nonterminal $n_d$ in a rule for $n_s$. We greedily find the set $\mathcal{N}$ of the minimal set of nodes that covers all the cycles in this graph. Then we make $D$ copies of every nonterminal $v^1,\hdots,v^d$ for all $v \in V$, and for every rule
\[
\scfgrule{v}{\rightarrow \sigma^1_0v_1\sigma^1_1\cdots v_n\sigma^1_n}{\sigma^2_0v_1\sigma^2_1\cdots v_n\sigma^2_n} 
\]
we replace it with $D$ copies of every rule where copy $d$ of a rule uses copy $d+1$ of a nonterminal $v_i$ if $v_i \in \mathcal{N}$:
\begin{eqnarray*}
\scfgrule{v^d}{\sigma^1_0d(v_1)\sigma^1_1\cdots d(v_n)\sigma^1_n}{\sigma^2_0d(v_1)\sigma^2_1\cdots d(v_n)\sigma^2_n}
\end{eqnarray*}
where $d(v_i)= v^{d+1}_i$ if $v_i \in \mathcal{N}$ and $v^{d}_i$ otherwise. For our experiments, we set $D=10$, which we find covers all the examples that we use.

Then, to select from a finite set of parses, we generate the corresponding program for each parse, and use a set of heuristic rules to discard programs that we know are incorrect. These rules include discarding programs that call \texttt{\form{Yield}} multiple times, as in \texttt{\form{(Yield :output (Yield :output ...))}}, and discarding programs that call \texttt{\form{CreatePreflightEventWrapper}} without calling \texttt{\form{CreateCommitEventWrapper}}. In practice we find that our heuristic rules can recover the correct parse 90\% of the time.

\subsection{Character-level Parsing}
\label{subsec:scfg-calflow:char}

When writing English templates, it would be inconvenient to ensure that the terminals of the grammar line up exactly with the tokens used by the language model.
Different LMs sometimes use subtly different tokenizers, and it would be especially inconvenient to write a different grammar for each LM.
In order to handle differences between the LM's tokenizer and the terminals of the grammar, we effectively treat the grammar as one whose terminals are all single characters.
Then to implement \constrainFunction, we advance each LM-proposed token character-by-character and return the set of tokens who, after being fully consumed, still have live Earley chart items.
By catering to the LM's preferred tokenization, we ensure that the LM's likelihood after incremental search matches the likelihood the LM would have assigned had it been given the full string to begin with.

\section{Intermediate Representation}
\begin{figure*}
\begin{center}
\includegraphics[scale=0.5]{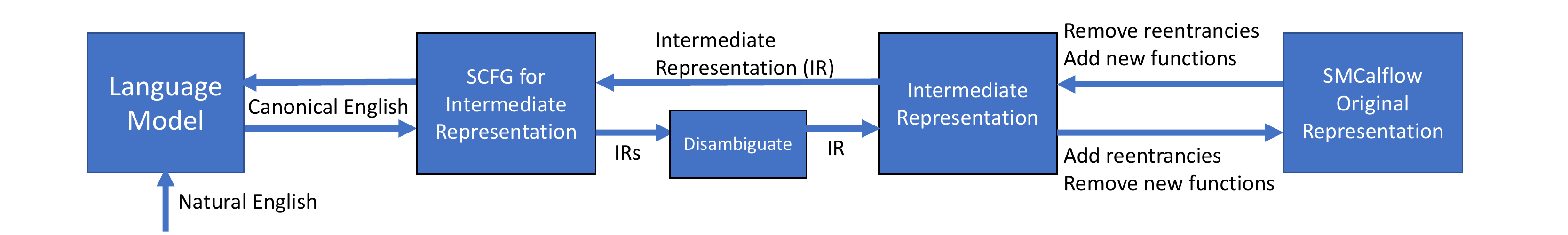}
\end{center}
\vspace{-1em}
\caption{Our pipeline for SMCalflow. We first convert SMCalflow's original representation to an intermediate representation, upon which we induce an SCFG. This SCFG is used to generate pairs of natural and canonical English utterances, which is used to train a language model to predict a canonical English utterance given a natural one. Predicted canonical English utterances are then mapped back into intermediate meaning representations, which can then be transformed back into the original representation.}
\label{fig:pipeline}
\end{figure*}
While we have described how to build an SCFG for mapping between meaning representations and canonical representations, we still have two problems. The first problem is that unfortunately as constructed, the SCFG cannot handle reentrancies, where expressions are cached in variables inside \texttt{let} expressions and then used multiple times. 
The second problem arises from the fact that it is impossible to engineer the English templates in a way that they produce natural English utterances for every possible composition of functions. For example, our English template for \texttt{get(\$object, \$path)} is \texttt{\$path of \$object}, which produces fluent English when getting the start time of an event, like in ``start time of event''. However, consider the program needed to deleting an event: \texttt{(DeleteCommitEventWrapper :event (DeletePreflightEventWrapper :id (get (constraint[Event]) \#(Path "id")))}. The SCFG would translate this program into ``delete id of event'' when we would prefer something closer to ``delete event.''

To solve both these problems, instead of inducing an SCFG based on the original SMCalflow representation, we instead first transform SMCalflow into an intermediate representation that 1) does not contain any reentrancies and 2) replaces common program fragments with calls to macros, and induce an SCFG on that the resulting intermediate representation. See Figure \ref{fig:pipeline} for a visualization of our entire process.

\subsection{Reentrancies} To remove reentrancies, given an expression of the form \texttt{(let (var binding) (body))} where the \texttt{body} contains usages of \texttt{var}, we replace the first usage of \texttt{var} (in postorder traversal) with \texttt{binding} and all other usages into calls to \texttt{(referWithinTurn T)} where $T \in \mathcal{T}$ is the return type of the \texttt{body} expression and \texttt{referWithinTurn} is a new function that retrieves the most ``salient'' evaluation of type $T$ from elsewhere in the program for the current utterance. 

Given a program $p$ in the intermediate representation, to convert a call to \texttt{(referWithinTurn T)} back into a \texttt{let} expression (to map from the intermediate representation back to the original), we find the first expression $e$ of type $T$ in $p$ (in postorder traversal), and replace $p$ with the \texttt{let} expression \texttt{(let (x $e$)  }sub\texttt{($p$, $e$, x))}, where sub replaces all expressions $e$ in $p$ with \texttt{x}. Note that this transformation is lossy. By picking the first expression that matches, it is possible to make mistakes, but we find in practice that such a heuristic is often good enough.

\subsection{Macros}
To reduce the number of unnatural sounding utterances produced by our grammar, we define macros to capture common program fragments, and then replace those fragments with calls to those macros. For example, we define a macro \texttt{DeleteWrapper(\$event)}, which we use to replace fragments that look like \texttt{(DeleteCommitEventWrapper :event (DeletePreflightEventWrapper :id (get \$event \#(Path "id")))}.
After defining macros, we add the macros to the set of functions $\mathcal{F}$ and corresponding English templates. In the case of \texttt{DeleteWrapper}, we write the template \texttt{delete \$event}.
In total, we define 15 new functions and find that they significantly help fluentize the resulting English. When translating from the intermediate representation back to the original SMCalflow representation, we can remove these new functions by simply replacing them with their definitions.

\section{Stratified Datasets}
\label{sec:stratified}

The motivation for our stratified datasets is to try and imitate what a small dataset over SMCalFlow or similar would look like had it been collected with a small target size in mind (i.e., collect and annotate 100 {\em representative} dialogue turns).
In this case, we expect that domain developers would do their best to guarantee that each supported SMCalFlow functionality appears in at least one example.
Such functionalities can be described by the functions that are used in the respective programs.
Therefore, our goal is to produce small subsets of the original large dataset ($\sim$120,000 dialogue turns), which guarantee that each supported function appears in at least $k$ examples ($k=1$ in our experiments).
The procedure we used to do this consists of three steps that we describe in the following paragraphs.

\paragraph{Function Histogram Extraction.}
We first extract function histograms for each example in our data.
This step consists of collecting all the function signatures that appear in each example and then constructing a histogram over these signatures for our train, validation, and test datasets.

\paragraph{Rare Functions Filtering.}
SMCalFlow contains some examples that use very rare functions.
These examples seem to be the result of annotation errors or incomplete data migrations and thus we do not want to include them in our stratified datasets.
Note that including them would also render complete coverage of the data (in terms of functionality) impossible with only 100 or 300 examples.
Therefore, in this step we:
(i) use the extracted histograms to collect all functions that appear less than $n$ times in the training data ($n=10$ in our experiments),
(ii) remove all examples that contain any of these functions across all of the train, validation, and test data, and 
(iii) update the dataset histograms to reflect the filtered data distributions.

\paragraph{Stratified Sampling.}
Our goal in this step is to use the function histograms and sample subsets of the filtered datasets which guarantee that each function appears in at least $k$ examples in each sample.
Let $m$ be the total number of examples in the dataset we are sampling from, and let $f$ be the total number of functions after the previous filtering step is applied.
We formulate our sampling problem as a mixed-integer program (MIP):
\begin{alignat}{2}
    \max_{\bm{x}}\quad & \bm{x}^\top \bm{c},       && \quad\textcolor{gray}{\textsc{Objective}} \\
    \textrm{s.t.}\quad & \bm{x}^\top\bm{1} = s,    && \quad\textcolor{gray}{\textsc{Target Size}} \\
                       & \bm{x}^\top\bm{H} \geq k, && \quad\textcolor{gray}{\textsc{Coverage}}
\end{alignat}
where $\bm{x} \in \{0, 1\}^m$ denotes whether or not an example is included in the subset we are sampling, $\bm{c} \in \mathbb{R}^m$ is a random vector sampled from the standard Gaussian distribution, $s$ is the target dataset size, and $\bm{H} \in \{0, 1\}^{m \times f}$ denotes the function membership for each example (i.e., $H_{ij} = 1$ specifies that example $i$ contains function $j$).
In our experiments we used the open-source JOpt solver, which can be found at \mbox{\url{https://github.com/blubin/jopt}}.

\section{Overnight}
\subsection{Our reproduction of Cao et al. (2019)}
\label{sec:caoetal-repro}

In order to investigate how a state-of-the-art method for Overnight performs when it is only given 200 training examples for each domain, we downloaded the code from \url{https://github.com/rhythmcao/semantic-parsing-dual} and made the following modifications:
\begin{itemize}
\item We used 200 training examples for each domain, the same examples as used in experiments with our methods.
\item Overnight does not have an official development set. Rather than holding out 20\% of the 200 training examples (the default methodology) as a development set to use for early stopping, we randomly sampled a development set with a size of 20\% of the original training set, from the original training set with the 200 chosen earlier excluded. 
\item We increased the total number of max epochs before stopping to 200, from 100. The code evaluates the model after each epoch on the development set, and chooses the snapshot that performed best on the development set.
\end{itemize}

\subsection{Miscellaneous details}
For our GPT-2, GPT-3, and BART experiments using meaning representations as the target output,
we removed all instances of the string \texttt{\small edu.stanford.nlp.sempre.overnight.SimpleWorld.} from the meaning representations, as it is redundant.
\section{Finetuning Experiments}
For our finetuning experiments, we use BART-large model which has 406 million parameters,
and the GPT2-XL model which has 1.5 billion parameters. We train each model using the causal LM loss for 20,000 steps, where we linearly warmup the learning rate for the 
first 1000 steps, and then reduce the learning rate by a 
factor of 0.999 every $t$ steps. For choosing hyperparameters, we perform a grid search by choosing the maximum learning rate from the set $\{10^{-5}, 10^{-6}\}$ and $t$ from the set $\{2, 4, 6, 8\}$. The best hyperparameters were chosen based on performance on a development set. We use a batch size of $32$, clip gradient norm at 10, and set a minimum learning rate threshold of $10^{-9}$.

\label{sec:gpt2}
We add some additional experimental results using finetuned GPT-2 XL in Tables \ref{tab:appendix_overnight} and \ref{tab:appendix_break}.  %
\begin{table*}
\centering
\notsotiny\frenchspacing
\setlength{\tabcolsep}{3pt}
\renewcommand{\arraystretch}{0.9}
\begin{tabularx}{\textwidth}{Xrrrrrrrrr}
\toprule
\textbf{Model} & \textbf{Train} \textit{n} & \textbf{Basketball} & \textbf{Blocks} & \textbf{Calendar} & \textbf{Housing} & \textbf{Publications} & \textbf{Recipes} & \textbf{Restaurants} & \textbf{Social} \\
\midrule
GPT-2$^f$ Constrained Canonical & 200 & 0.836 & 0.549 & 0.804 & 0.640 & 0.752 & 0.787 & 0.762 & 0.726 \\
GPT-2$^f$ Constrained Meaning & 200 & 0.831 & 0.516 & 0.732 & 0.677 & 0.727 & 0.778 & 0.768 & 0.671 \\
GPT-2$^f$ Unconstrained Canonical & 200 & 0.798 & 0.509 & 0.762 & 0.603 & 0.720 & 0.745 & 0.705 & 0.632 \\
GPT-2$^f$ Unconstrained Meaning & 200 & 0.821 & 0.506 & 0.708 & 0.646 & 0.671 & 0.755 & 0.753 & 0.626 \\
\bottomrule
\end{tabularx}
\caption{Accuracy on Overnight dataset using GPT-2 XL.}
\label{tab:appendix_overnight}
\end{table*}

\begin{table}[H]
\centering
\notsotiny\frenchspacing
\setlength{\tabcolsep}{3pt}
\renewcommand{\arraystretch}{0.9}
\begin{tabularx}{\columnwidth}{Xrl}
\toprule
{\bf Model} &$n$  & {\bf nem}   \\ %
\midrule
Coleman \& Reneau& 44,321 & 0.42	\\
\newcite{Wolfson2020Break}& 44,321&0.29\\
Arad \& Sapir& 44,321&0.16 \\
BART$^f$ Unconstrained Meaning & 200 & 0.10 \\
\midrule
GPT-2$^f$ Constrained Canonical & 200 & 0.18 \\
GPT-2$^f$ Constrained Meaning & 200 & 0.17 \\
GPT-2$^f$ Unconstrained Canonical & 200 & 0.13 \\
GPT-2$^f$ Unconstrained Meaning & 200 & 0.13 \\
\bottomrule
\end{tabularx}
\caption{NEM accuracy on the Break dataset using GPT-2 XL.}
\label{tab:appendix_break}
\end{table}

\section{Computing Infrastructure}
For the GPT-3 experiments, we used OpenAI's GPT-3 API hosted on Microsoft Azure. For the finetuning experiments, we used NVIDIA DGX-2 machines contaning NVIDIA Tesla V100 GPUs.
\section{Further Discussion}
\label{sec:discussion}

A common thread among all of our datasets, and arguably semantic parsing in general, is that annotation subtleties cause problems for automated methods and annotators alike. 
For example, on the Calendar subset of Overnight, we found that of our best model's 18 errors,
8 were legitimately wrong, 7 were annotation errors that the model actually got right, and 3 differed only by equality strictness -- which is often left ambiguous in natural language. For example, for the input: \textit{tell me the all meetings begins after 10am or 3pm},
the annotated canonical form in the data is: \controlled{meeting whose start time is at least 10am or 3pm}, but our system predicted: \controlled{meeting whose start time is larger than 10am or 3pm}.%
We would expect low interannotator agreement on this subtle distinction ($\geq$ vs.\@ $>$), just as we would expect GPT-3 to perform poorly.
As another example, on the Calendar subdomain of Overnight, our best model's denotation accuracy @$K$ saturated at 0.98 when $K \ge 5$;
but we found that the 2 remaining errors were caused by annotation mistakes on utterances that the model correctly interpreted.

\end{document}